\documentclass{article}

\usepackage[english]{babel}

\usepackage[utf8]{inputenc}
\usepackage{mathtools}
\usepackage[letterpaper,top=2cm,bottom=2cm,left=2cm,right=2cm,marginparwidth=1.75cm]{geometry}

\usepackage{authblk}

\usepackage{amsmath}
\usepackage{amssymb}
\usepackage{amsfonts}
\usepackage{amscd}
\usepackage{amsthm}
\usepackage{subcaption}

\usepackage{booktabs}

\theoremstyle{definition}

\theoremstyle{definition}
\newtheorem{example}{Example}[section]

\usepackage{multirow}
\usepackage{array}
\usepackage{graphicx}
\usepackage[colorlinks=true, allcolors=blue]{hyperref}

\title{A MULTISTEP  SEGMENTATION ALGORITHM FOR VESSEL EXTRACTION IN MEDICAL IMAGING}

\author[1]{Nasser Aghazadeh}
\author[2]{Ladan Sharafyan Cigaroudy}
\affil[1,2]{Department of Applied Mathematics, Azarbaijan Shahid Madani University, Tabriz, Iran\newline emails: aghazadeh@azaruniv.ac.ir$^1$,~ sharafyan@azaruniv.ac.ir$^2$}

\date{}
\begin{document}
\maketitle

\begin{abstract}
The main contribution of this paper is to propose an iterative procedure for tubular structure segmentation of 2D images, which combines tight frame of Curvelet transforms with a SURE technique thresholding which is based on
principle obtained by minimizing Stein Unbiased Risk Estimate for denoising. This proposed algorithm is mainly based on the TFA proposal presented in \cite{Cai1, Cai2}, which we use eigenvectors of Hessian matrix of image for improving this
iterative part in segmenting unclear and narrow vessels and filling the gap between separate pieces of detected vessels. The experimental results are presented to demonstrate the effectiveness of the proposed model. \bigskip

 \bigskip

\end{abstract}

\section{Introduction}
Segmentation is a useful tool with various applications in different fields such as medical diagnosis, digital system designing and spatial images. Distinguishing objects, homogeneous tissues, edges and ridges in images with mathematical tools is a branch of image processing \cite{Gonzalez}. Image segmentation is applied in medical sciences to distinguish
special tissue like tumors in tomography, ridges for vessel extraction in angiography and etc. The extraction of tubular
structure of vessels has fundamental role in some automated medical diagnosis.

The paper considers tubular structure segmentation that can be applied to vessel extraction. Depending on the purpose of
segmentation, various algorithms have been proposed to solve the segmentation problem. Among them there are two
main methods: based on edge \cite{joes} and based on the region \cite{A var}, both of which are used in our algorithm. We use Curvelet
which has important role in representing edges. Also we use gradient for pre-segmenting of pixels. Both of them are
edge detection tools. On the other hand, the region based technique which we use are the value of intensity function and
eigenvectors of Hessian matrix for the pixels belonging to a subdomain of image which has potentiality of being vessel.

At the beginning of our algorithm we compute the discrete Gradient of image for At the beginning of our algorithm we compute the discrete Gradient of image for dividing pixels into two regions: one
region consists of vessel pixels and the second includes the pixels in the background. In addition, we need the
eigenvectors of Hessian matrix because of their similar directions in the same and near region. In section 2, we explain
the method of computing first and second order derivatives of the image which are used in the discrete Gradient of
image and Hessian matrix. In section 3, we briefly introduce the tight frame of Curvelets transform, that we choose for
decomposition and reconstruction of the image to arrange denoising part. After the image decomposition (Curvelet transform of image) and before the reconstruction (inverse of Curvelt transform), we use SURE technique thresholding \cite{Insight} for denoising. In section 4, we describe this technique. In section 5, we completely illustrate our iterative algorithm
step by step, all calculated parameters of previous sections are used in this part. At last, in section 6, we present four
examples to show our experimental results.
\section{Gaussian gradient and eigenvector of Hessian matrix}
In the angiography images, ridges are natural indicators of vessels. Ridges are defined as points where the image has an
extremum in the direction of the largest surface curvature. Thus, the first derivative of the intensity in the direction of the
largest surface curvature changes sign. The interested reader can consult \cite{machinevision} and the references therein for a more detailed review of the surface curvature.
The direction of largest surface curvature is the eigenvector $ v $ of the matrix of second order derivatives of the image corresponding to the largest absolute eigenvalue $\lambda $. This matrix is referred to as the Hessian matrix $ H$ for any 2D image $ f $

\begin{center}
$\nabla^2f(X)= H(X)=\begin{bmatrix}  f_{xx}&f_{xy}\\f_{yx}&f_{yy}  \end{bmatrix} . $
\end{center}
 Taking derivatives of discrete images is an ill-posed operation \cite{Image}, thus they are taken at a scale $ \sigma $ using the Gaussian scale-space technique\cite{Im.St}. The main idea is that the image derivatives can be taken by convolving the image with derivatives of a Gaussian kernel with respect to a coordinate of image such as $ x $,
\begin{equation}
\label{Gaussian}
\frac{\partial f(X,\sigma)}{\partial x}=\frac{1}{2\pi\sigma^2}\int_{X^{\prime}\in R^2}\frac{\partial e^{-\Vert X-X^{\prime}\Vert^2/2\sigma^2}}{\partial x}f(X^{\prime})dX^{\prime} ,
\end{equation}
where $ X=(x,y) $. The derivation with respect to $y$ is done in the same manner. Mixed and higher order derivatives are computed by taking mixed and higher order derivatives of the Gaussian kernel.

Another idea that we use in the steps of our algorithm is the fact that the directions of eigenvectors ($ v $) of the ridge pixels should be similar \cite{joes}. If the pixels have similar orientation the scalar product will be close to 1. One parameter that we calculate and use in each step is the mean value ($ mean^{(i)} $) of scalar product of eigenvector $ v $ of the pixel of maximum gradient with the  other pixels in $ \Lambda^{(i)} $ (in each step, $ \Lambda^{(i)} $ includes the pixels which have potential of being vessel).
\section{Tight frame of Curvelets transform}
Despite the fact that wavelets have had a wide impact in image processing (see \cite{ten,discovering}) they fail to efficiently
represent objects with highly anisotropic elements such as lines or curvilinear structures (e.g. edges).
The reason is that wavelets are non-geometrical and do not exploit the regularity of the edge curve.
The Ridgelet and the Curvelet transforms were developed as an answer to the weakness of
the separable wavelet transform in sparsely representing of what appears to be simple building atoms
in an image, e.g. lines, curves and edges. Curvelets and Ridgelets take the form of basis elements
which exhibit high directional sensitivity and are highly anisotropic \cite{Curv1,Curv2}.
Ridgelets and Curvelets are special
members of the family of multiscale orientation-selective transforms, which have recently led to a flurry of research
activity in the field of computational and applied harmonic analysis. Curvelet transform developed by Candes and
Donoho is a multiscale transform designed to represent edges and other singularities along curves much more efficiently
than the traditional transforms, i.e., using fewer coefficients for a given accuracy of reconstruction \cite{Insight}.

In this paper, we use Fast Discrete Curvelet Transforms from \cite{Candes2006}. This transformation is based on unequally-spaced
fast Fourier transforms (USFFT). This implementation is fast in the sense that it runs in $O(n^2 \log{n})$ flops for $n$ by $n$
Cartesian arrays; in addition, it is invertible, with rapid inversion algorithms of about the same complexity. We use
MATLAB based CurveLab toolbox and the software CurveLab, which implement this transformform from \cite{Candes2006}.
If we introduce $ C(\cdot) $ as Curvelet transform (decomposition) operator, $ T_{\lambda} $  as thresholding operator and $ IC(\cdot) $ as inverse Curvelet transform (reconstruction) operator, then we use
\begin{equation}
\label{curve}
IC(T_{\lambda}(C(f^{(i)})))
\end{equation}
for denoising $ f^{(i)} $ in each step. It is noticeable that we have $ IC(C(f))=f $, because  we use tight frame Curvelets \cite{frame, Cai12} .
\section{Thresholding}We use a thresholding technique based on a principle derived by minimizing Stein Unbiased Risk Estimate (SURE) \cite{Insight}.
A standard application of SURE is to choose a parametric form for an estimator, and then optimize the values of the
parameters to minimize the risk estimate. This technique has been applied in several settings. It has been used by
Donoho and Johnstone to determine the optimal shrinkage factor in a wavelet denoising setting \cite{Donoho1995}. We use this
technique for estimate the hard thresholding parameter for denoising based on Curvelet as follow.

We change $ Nx\times Ny$ matrix of image $ f $ into vector $ F$ with size $ Nxy=Nx*Ny $ and sort its second powers:
\[a=sort\lbrace \vert F_j \vert^2\rbrace_{j=1}^{Nxy},\]
and calculate other parameters:
\[b=\lbrace \Sigma_{r=1}^j a_r\rbrace_{j=1}^{Nxy}, \]
\[c=\lbrace Nxy-j\rbrace_{j=1}^{Nxy},\]
\[s=\lbrace b_j+c_ja_j\rbrace_{j=1}^{Nxy}, \]
\[risk=\left\{ \frac{n-2j+s_j}{n}\right\}_{j=1}^{Nxy},\]
then let
\[risk_{ibest}=min\lbrace risk_j\rbrace_{j=1}^{Nxy},\]
and
\[\lambda_{thresh}=\sqrt{a_{ibest}} .\]
We use  $\lambda_{thresh}$ for image hard thresholding.
\section{The iterative algorithm}
The main idea of this algorithm is based on TFA \cite{Cai1,Cai11}. We explain the algorithm with more details that we have added to it.
At first, we define $ f^{(i)}\equiv f $ for $ i=0 $ and \\
\begin{equation}
\label{Lambda0}
\Lambda^{(i)} \equiv\lbrace(x,y)\in\Omega \;\vert\; \Vert(\bigtriangledown f )_{(x,y)}\Vert_2\geq\epsilon\rbrace ,
\end{equation}
for a small supposed $ \epsilon $.
Here $ (\bigtriangledown f )_{(x,y)}$ is the discrete gradient of $ f $ at the pixel $ (x,y) $ that we compute with Gaussian scale-space technique. Now we continue the iterative process with the initial data $ \Lambda^0 $  and $f^0 $. \bigskip

\noindent {\bf STEP1}:
We compute the maximum gradient and find the pixel with maximum gradient  \\
\begin{equation}
\label{maxg}
 maxg=\max \lbrace \Vert\nabla f(x,y)\Vert_2 \; \vert\; (x,y)\in\Lambda^{(i)} \rbrace=\Vert\nabla f(\bar{x},\bar{y})\Vert_2, \hspace*{3mm}     (\bar{x},\bar{y})\in \Lambda^{(i)}
\end{equation}
\begin{equation}
\label{maxp}
maxp=(\bar{x},\bar{y}).
\end{equation}
\noindent {\bf STEP2}:
Compute four decision parameters, $ mean^{(i)} , M^{(i)}, Mn^{(i)}$ and  $Mp^{(i)} $:
\begin{equation}
\label{mean}
mean^{(i)}=\frac{\Sigma_{(x,y)\in \Lambda^{(i)}} v(maxp).v(x,y)}{\vert\Lambda^{(i)}\vert},
\end{equation}
and
\begin{equation}
\label{M}
M^{(i)}=\frac{\Sigma_{(x,y)\in\Lambda^{(i)}}f^{(i)}(x,y)}{\vert\Lambda^{(i)}\vert},
\end{equation}
and
\begin{equation}
\label{Mp}
Mp^{(i)}= \frac{\Sigma_{\lbrace (x,y)\in\Lambda^{(i)}\vert f^{(i)}(x,y)\geq M^{(i)}\rbrace}f^{(i)}(x,y)}{\vert\lbrace (x,y)\in\Lambda^{(i)}\;\vert\; f^{(i)}(x,y)\geq M^{(i)}\rbrace\vert},
\end{equation}
and
\begin{equation}
\label{Mn}
Mn^{(i)}= \frac{\Sigma_{\lbrace (x,y)\in\Lambda^{(i)}\vert f^{(i)}(x,y)\leq M^{(i)}\rbrace}f^{(i)}(x,y)}{\vert\lbrace (x,y)\in\Lambda^{(i)}\;\vert\; f^{(i)}(x,y)\leq M^{(i)}\rbrace\vert}.
\end{equation}
We define
\begin{equation}
\label{a,b}
\alpha_i\equiv \max \left\{\frac{M^{(i)}+Mn^{(i)}}{2},0 \right\}, \quad \beta_i\equiv \min \left\{\frac{M^{(i)}+Mp^{(i)}}{2}, 1 \right\}.
\end{equation}
The main difference of our algorithm and TFA in this step is equation (\ref{mean}). \bigskip

\noindent{ \bf STEP3}:
Image can be tresholded by introducing the $ f_{t}^{(i)} $ which  subdivide the image into  three parts:
\begin{equation}
\label{ftresh}
f_{t}^{(i)}(x,y)=\left \{
\begin{array}{ccc}
0&\,\,\,  f^{(i)}(x,y)\leq \alpha_i \\
1 &\,\,\, if\{f^{(i)}(x,y) \geq M^{(i)},\quad \text{and} \quad  v(maxp).v(x,y) \geq mean^{(i)}\}\\
\, & or\, f^{(i)}(x,y)\geq \beta_i \\
\frac{f^{(i)}(x,y)-min_i}{max_i-min_i} & \,\,\,o.w.
\end{array}
\right.
\end{equation}
where
\[max_i=\max\lbrace f^{(i)}(x,y)\vert  \alpha_i\leq f^{(i)}(x,y) \leq\beta_i,(x,y)\in \Lambda^{(i)}\rbrace,\]
\[min_i=\min\lbrace f^{(i)}(x,y)\vert  \alpha_i\leq f^{(i)}(x,y) \leq\beta_i,(x,y)\in \Lambda^{(i)}\rbrace.\]
We use $ mean^{(i)} $ for second constraint of $ f_t^{(i)} $.
The set of the remaining pixels that should be classified is denoted by:
\begin{equation}
\label{Lambda}
\Lambda^{(i+1)}=\lbrace(x,y)\;\vert \;0<f_t^{(i)}(x,y)<1, (x,y)\in\Omega   \rbrace .
\end{equation}
In the next step, we denoise and smooth $ f_t^{(i)} $ on $ \Lambda^{(i+1)} $ with Curvelets based algorithm.\bigskip

\noindent {\bf STEP4} (denoising and smoothing):
In this step we apply equation (\ref{curve}) to $ f_t^{(i)} $ on  $ \Lambda^{(i+1)} $ to get $f^{(i+1)}$. For the other pixels  which belong to $ \Omega-\Lambda^{(i+1)} $,  we have $f^{(i+1)}= f_t^{(i)} $. To write it out clearly, let $ F_t^{(i)}$ $=vec(f_t^{(i)}) $ where $ vec(.) $ represents the vector form of a two-dimensional (2D) pixel-array (size $ Nx\times Ny$ ) by concatenation in the usual columnvise fashion, and $ P^{(i+1)} $ be the diagonal matrix where the diagonal entry is 1 if the corresponding index of pixel be in  $ \Lambda^{(i+1)} $, and 0 otherwise. Then
\begin{equation}
\label{f(i+1)}
F^{(i+1)}\equiv (I-P^{(i+1)})F_t^{(i)}+P^{(i+1)} vec( IC(T_{\lambda}(C(f_t^{(i)}))).
\end{equation}
By reordering the entries of the vector $F^{(i+1)}$ into columns, we obtain the denoised and smoothed image $f^{(i+1)}$ \cite{Cai2}.
In every iteration the cost of (\ref{f(i+1)}) is reduced, because it depends on the pixels of $ \Lambda^{(i+1)} $, that decreases in each iteration. When $  \Lambda^{(i+1)}=\emptyset $, the iteration terminates, and it happens immediately after the values of $ f_t^{(i)} $ are just 1 and 0. The pixels with value 0 are considered as background and the pixels with value 1 constitute the vessels.  Here, we summarize the algorithm:

\begin{enumerate}
  \item[$\bullet$]  Input: given image $ f $.
  \item[$\bullet$] Set $ f^{(0)}=f $ and $ \Lambda^{(0)} $ by (\ref{Lambda0}).
  \item[$\bullet$] Do $i=0,1,2,...$
  \begin{enumerate}
    \item[-] Compute $ maxg $ by (\ref{maxg}) and $ maxp $ by (\ref{maxp}).
    \item[-] Compute $ mean^{(i)} $ by (\ref{mean}) and $ [ \alpha_i,\beta_i] $ by (\ref{a,b}).
    \item[-] Threshold $f^{(i)} $ into $ f_t^{(i)}$ by (\ref{ftresh}).
    \item[-] Compute $\Lambda^{(i+1)} $  by (\ref{Lambda}).
    \item[-] Stop if  $\Lambda^{(i+1)}=\emptyset $, therefore $ f_t^{(i)}$ is a binary image.
    \item[-] Compute $f^{(i+1)} $ by $ f_t^{(i)}$ by (\ref{f(i+1)}).
  \end{enumerate}
  \item[$\bullet$] Output: binary image $ f_t^{(i)}$.
\end{enumerate}

\newtheorem{mydef}{Theorem}
\begin{mydef}
The algorithm converges to a binary image within a finite number of steps.
\end{mydef}
\begin{proof}
The proof follows the same arguments as in \cite[Theorem 3.1]{Cai1}.
Since  $\vert\Lambda^{(i+1)} \vert\leq\vert\Lambda^{(i)}\vert $ and
$ \vert\Lambda^{(0)}\vert $ is finite, thus finite steps leads to
$ \vert\Lambda^{(i)}\vert=0 $ and $ f^{(i)}_t $becomes binary.
\end{proof}

 \section{Experimental results}

\begin{figure}[h] 
\center
\begin{tabular} {ccc}
\hspace{-1.0cm}
\includegraphics[height=1.5in,width=1.5in]{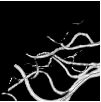} \hspace{1.0cm}
\includegraphics[height=1.5in,width=1.5in]{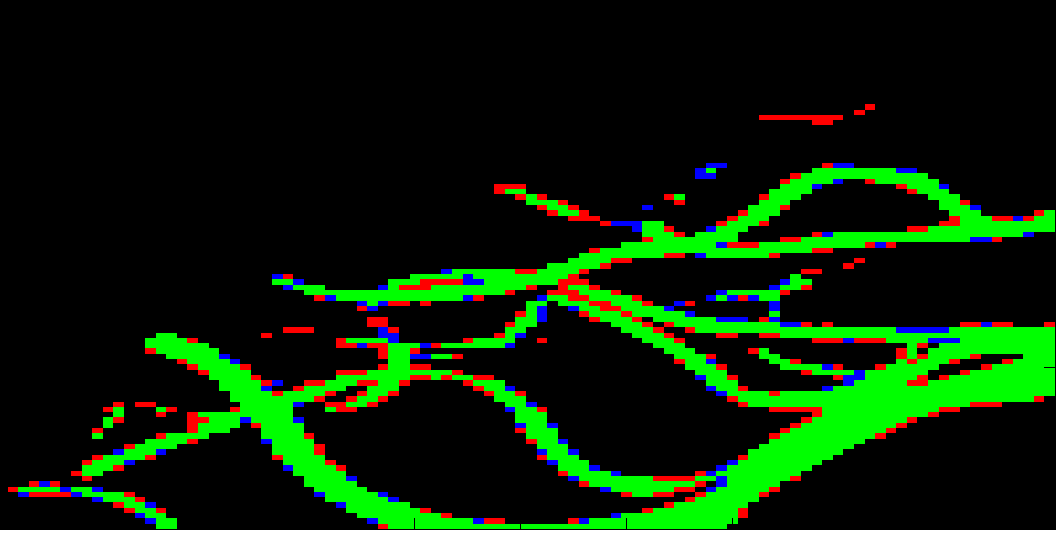} \hspace{1.0cm}
\includegraphics[height=1.5in,width=1.5in]{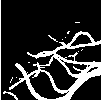} \hspace{1.0cm}
 \end{tabular}
\vspace{0.5cm}
\caption{True image (left),  the result of the first step of our algorithm (center), the blue pixels are those that our algorithm distinguished them as vessels. The value of $f_t$ in red pixels is  between 0 and 1 which must categorized into vessel or background in the next steps.  The final segmented image (right)}
\label{FIG01}
\end{figure}

 In this section, we present some medical images. We test our proposed algorithm and compare it with TFA \cite{Cai1,Cai11}, and a Mamford-Shah based algorithm that Yibao Li and Junseok Kim used for biomodal image segmentation \cite{YJ} (We refer to it by LK in the sequel). For the color images, we choose green channel for segmentation, because of its high contrast in edges. For simplicity, we refer to our segmentation method as TFAE (TFA with $ eigenvector $). 
All computations are implemented in MATLAB 2011a running on an Intel Core i5 with about 16 significant decimal digits, and the codes of
methods are provided by the authors, except for tight frame transform part and in our examples, we fix  $ \sigma=2 $, and $ \epsilon=0.02 $.
 \begin{example}[Effect of eigenvectors to reduce the number of iterations]
In this example, we focus on the effect of scalar product of eigenvectors on reducing the number of iterrations. distinguishing vessels. We work on a small part of MRTA image with size $100\times 100$ (Figure \ref{FIG01}). The first step result shows that there are some pixels (blue pixels) that TFAE at the first step can segment them as vessels while the TFA algorithm can not do it at that step. In spite of the fact that the computation order
of our algorithm in each step is more than TFA's, TFAE reduces the number of iterations, therefore the runnig time of
TFAE does not change significantly.

\label{exam1}
\end{example}

\begin{figure}
\centering
  \begin{tabular}{cccc}
\begin{subfigure}[b]{0.22\textwidth}
   \hspace{-1cm}\includegraphics[width=\textwidth]{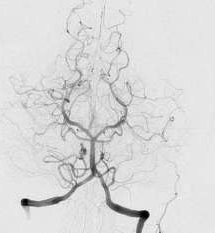}
                \caption{True}
                \label{fig:mm}
        \end{subfigure}
        \quad
        \begin{subfigure}[b]{0.22\textwidth}
                \includegraphics[width=\textwidth]{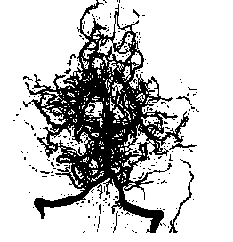}
                \caption{LK}
                \label{fig:mm_LK}
        \end{subfigure}
     \quad
           \begin{subfigure}[b]{0.22\textwidth}
                \includegraphics[width=\textwidth]{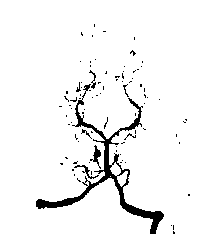}
                \caption{TFA}
                \label{fig:mm_cai}
        \end{subfigure}
\quad
          \begin{subfigure}[b]{0.22\textwidth}
                \includegraphics[width=\textwidth]{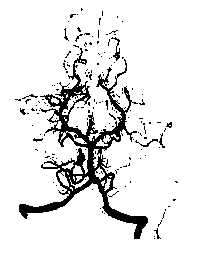}
                \caption{TFAE}
                \label{fig:mm_lad}
        \end{subfigure}\\
\begin{subfigure}[b]{0.22\textwidth}
  \hspace{-1cm}  \includegraphics[width=\textwidth]{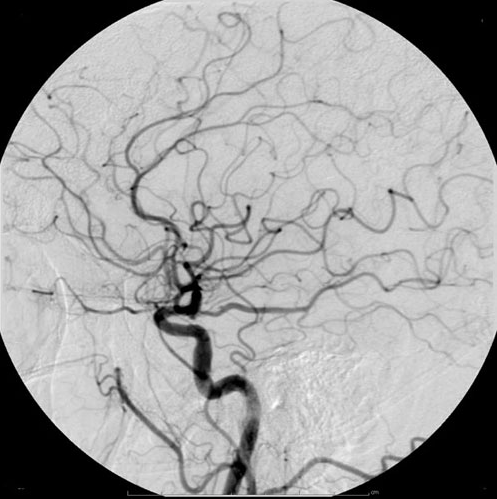}
                \caption{True}
                \label{fig:aghio4}
        \end{subfigure}
         \quad
         \begin{subfigure}[b]{0.22\textwidth}
                \includegraphics[width=\textwidth]{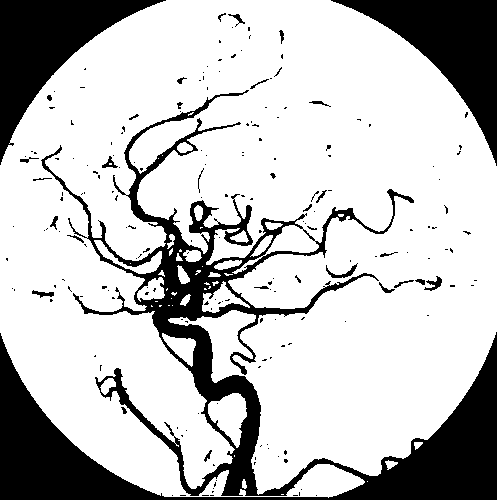}
                \caption{LK}
                \label{fig:aghio4_LK}
        \end{subfigure}
        \quad
         \begin{subfigure}[b]{0.22\textwidth}
                \includegraphics[width=\textwidth]{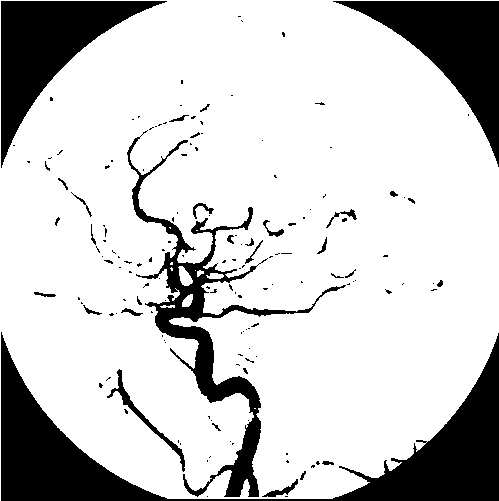}
                \caption{TFA}
                \label{fig:aghio4_cai_8it}
        \end{subfigure}
        \quad
         \begin{subfigure}[b]{0.22\textwidth}
                \includegraphics[width=\textwidth]{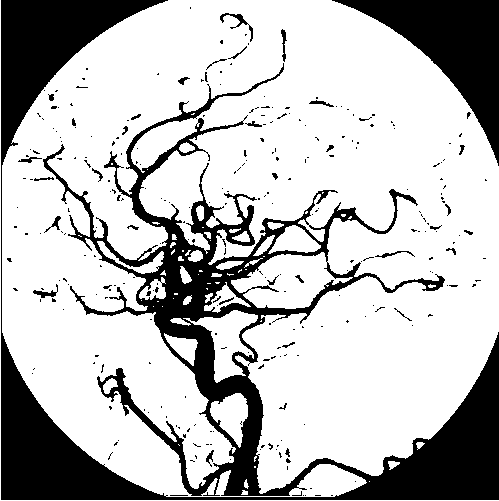}
                \caption{TFAE}
                \label{fig:aghio4_lad_4it}
        \end{subfigure}
        
     \end{tabular}
        \caption{True images are in the first column((a),(e)) , the results of LK are in the second column ( (b),(f)), the results of
TFA are in the third column ((c),(g)) and the results of TFAE are in the last column ( (d), (h)).}
\label{FIG02}
\end{figure}

\begin{figure}
        \centering
        \begin{subfigure}[b]{0.3\textwidth}
                \includegraphics[width=\textwidth]{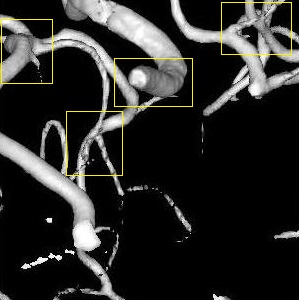}
                \caption{given}
                \label{fig:mrt-angio300y}
        \end{subfigure}
         \quad
       \quad
        \begin{subfigure}[b]{0.3\textwidth}
                \includegraphics[width=\textwidth]{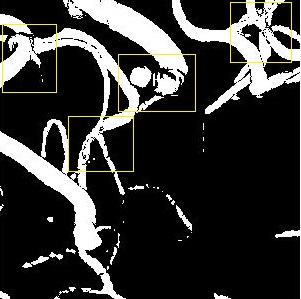}
                \caption{TFA}
                \label{fig:mrt-angio300-cai-curvey}
        \end{subfigure}
        \quad
         \begin{subfigure}[b]{0.3\textwidth}
                \includegraphics[width=\textwidth]{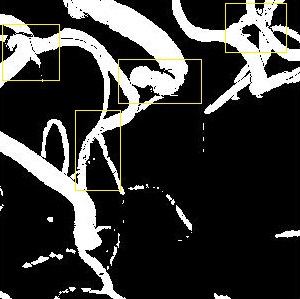}
                \caption{our algorithm}
                \label{fig:mrt-angio300-ladan-curvey}
        \end{subfigure}
         \caption{(a) True image. We draw yellow rectangular boxes around the pixels in the junction of multi vessels, or in the curvature of one vessel, which have darker intensity. (b) TFA segmentation result, (c) our algorithm segmentation result. }
\label{FIG03}
\end{figure}

\begin{figure}
      \centering
\begin{tabular}{cccc}
    \hspace{-1cm} \begin{subfigure}[b]{0.22\textwidth}
                \includegraphics[width=\textwidth]{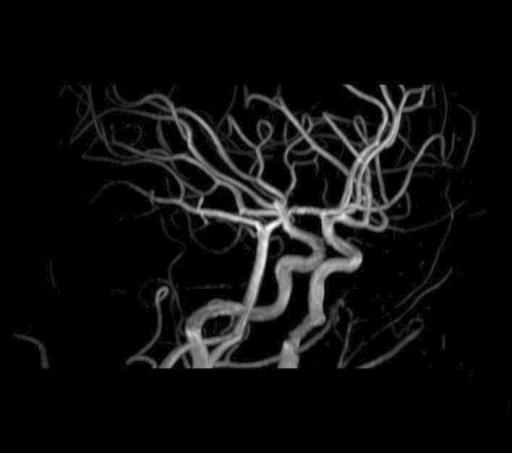}
                \caption{True}
                \label{fig:willistof10}
        \end{subfigure}
        \quad
         \begin{subfigure}[b]{0.22\textwidth}
                \includegraphics[width=\textwidth]{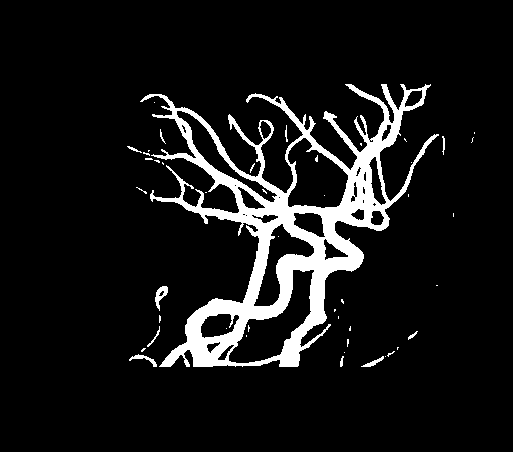}
                \caption{TFA}
                \label{fig:willistof10_cai}
        \end{subfigure}
        \quad
         \begin{subfigure}[b]{0.22\textwidth}
                \includegraphics[width=\textwidth]{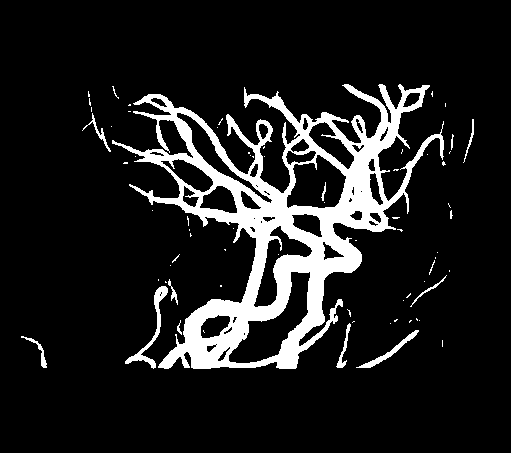}
                \caption{TFAE}
                \label{fig:willistof10_lad}
        \end{subfigure}
        \quad
         \begin{subfigure}[b]{0.22\textwidth}
                \includegraphics[width=\textwidth]{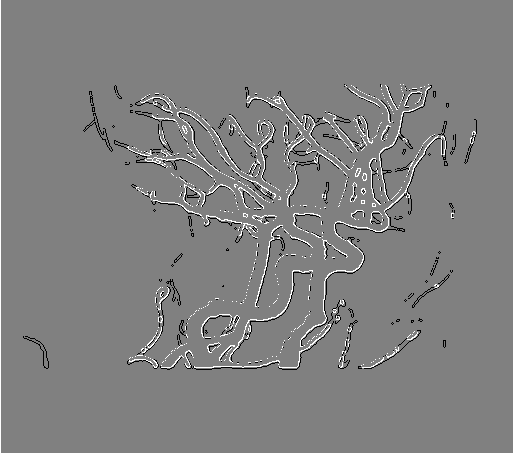}
                \caption{}
                \label{fig:willistof10_T}
        \end{subfigure}\\
       \hspace{-1cm}\begin{subfigure}[b]{0.22\textwidth}
                \includegraphics[width=\textwidth]{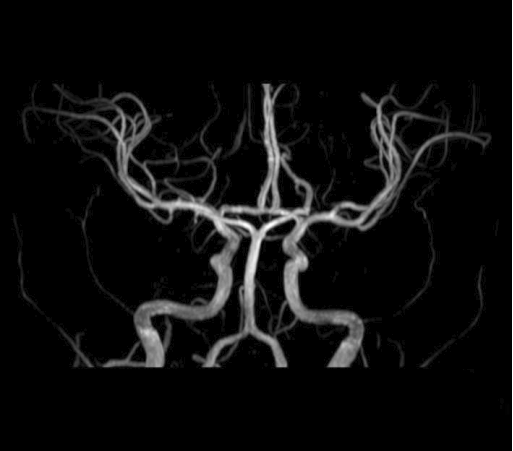}
                \caption{True}
                \label{fig:willistof30}
        \end{subfigure}
        \quad
         \begin{subfigure}[b]{0.22\textwidth}
                \includegraphics[width=\textwidth]{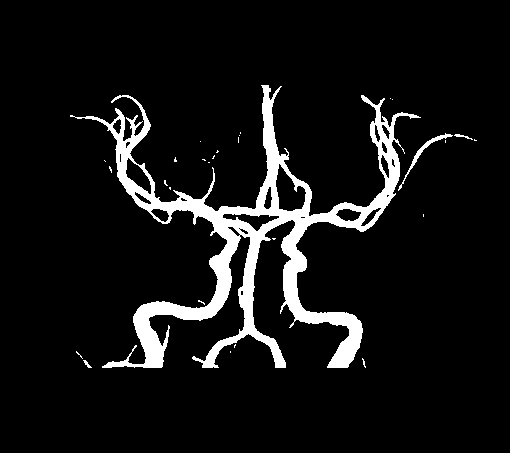}
                \caption{TFA}
                \label{fig:willistof30_cai}
        \end{subfigure}
        \quad
         \begin{subfigure}[b]{0.22\textwidth}
                \includegraphics[width=\textwidth]{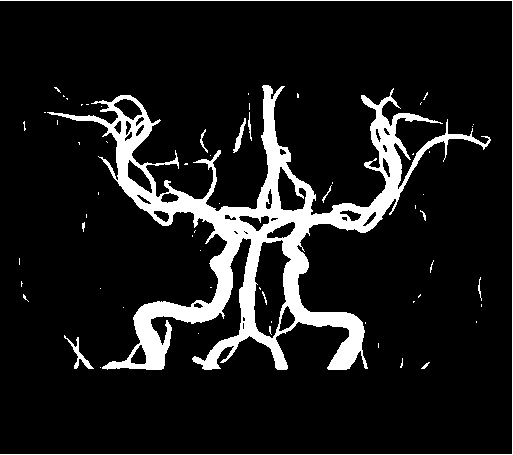}
                \caption{TFAE}
                \label{fig:willistof30_lad}
        \end{subfigure}
        \quad
         \begin{subfigure}[b]{0.22\textwidth}
                \includegraphics[width=\textwidth]{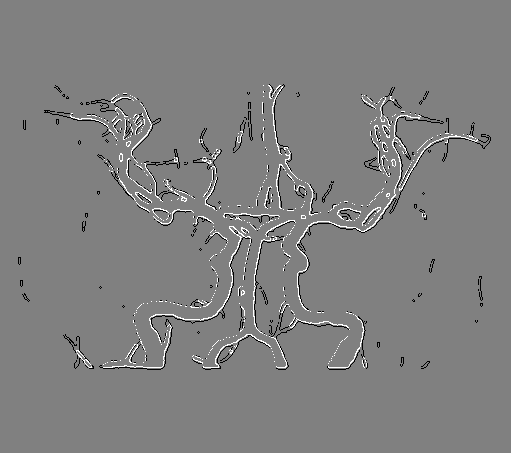}
                \caption{}
                \label{fig:willistof30_T}
        \end{subfigure}\\
                    \hspace{-1cm}\begin{subfigure}[b]{0.22\textwidth}
                \includegraphics[width=\textwidth]{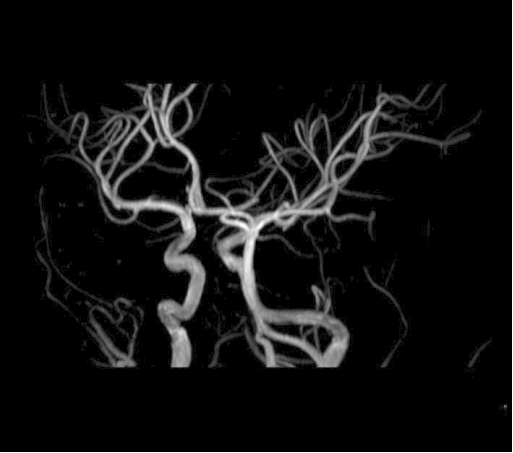}
                \caption{True}
                \label{fig:willistof45}
        \end{subfigure}
        \quad
         \begin{subfigure}[b]{0.22\textwidth}
                \includegraphics[width=\textwidth]{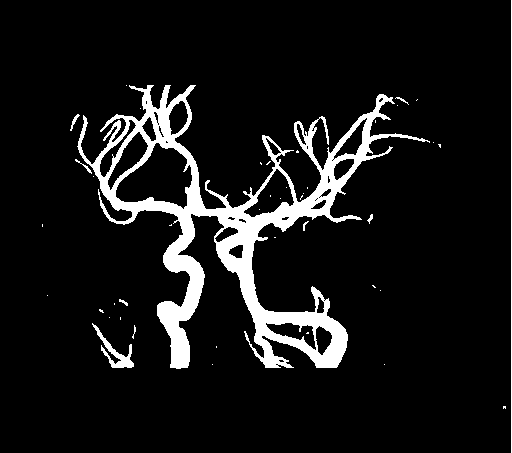}
                \caption{TFA}
                \label{fig:willistof45_cai}
        \end{subfigure}
        \quad
         \begin{subfigure}[b]{0.22\textwidth}
                \includegraphics[width=\textwidth]{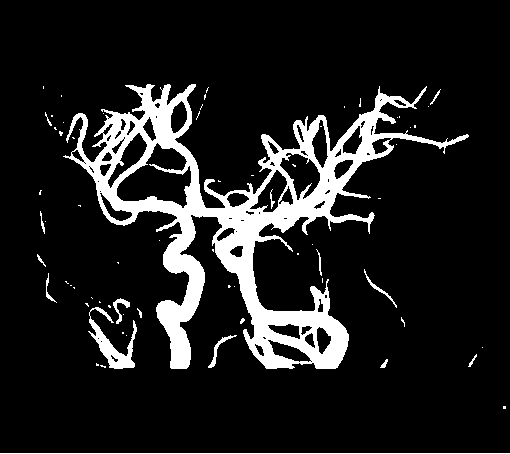}
                \caption{TFAE}
                \label{fig:willistof45_lad}
        \end{subfigure}
       \quad
         \begin{subfigure}[b]{0.22\textwidth}
                \includegraphics[width=\textwidth]{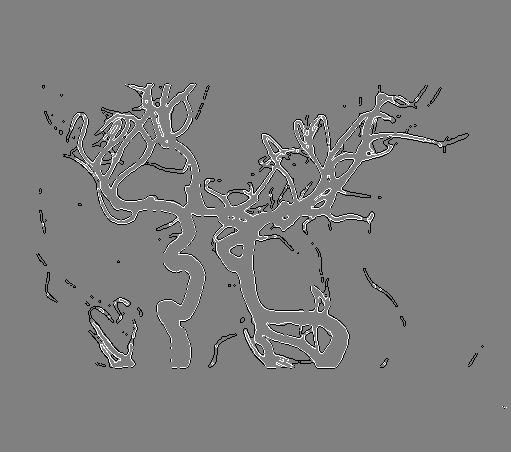}
                \caption{TFAE}
                \label{fig:willistof45_lad}
        \end{subfigure}
\end{tabular}         
         \caption{True images are in the first column, results of TFA algorithm are in the second column(b) ,the results of TFAE are in the third column(c) and the images of the forth column are the different between TFA and TFAE segmentation methods, the black pixels, which commonly belong to narrow and unclear vessels. }
\label{FIG04}
\end{figure}

 \begin{example}
 In this example we test two angiography $ 216\times 233 $  and $ 498\times 500 $ JPEG images (Figure \ref{FIG02}) by TFAE, TFA \cite{Cai1,Cai2} and LK \cite{YJ} methods. The results demonstrate that TFAE segments more unclear and narrow vessels specially in the end of vessels (the comparison between Figures \ref{fig:mm_lad} , \ref{fig:aghio4_lad_4it} with Figures \ref{fig:aghio4_cai_8it}, \ref{fig:aghio4_LK} and \ref{fig:mm_cai}), and less backgruond pixels as a vessel (Figure \ref{fig:mm_LK}).

\label{exam2}
\end{example}
\begin{example}[The dark vessel pixels]
In this example, we show that some vessel pixels with relatively dark intensity, can be distinguished as vessel by TFAE more than TFA (Figure \ref{FIG03}). In spite of difference between the intensity of this kind of pixels  and other vessel pixels, the direction of their eigenvector of Hessian matrix is similar and this is the point that they can be categorized as vessel by TFAE. For this aim, we choose a $ 300\times 300 $ MRT Angiography image  and provide the results in Figure \ref{FIG03}.
\label{exam3}
\end{example}
\begin{example}

3D time of flight MR angiography (3D-TOF-MRA) of the circle of Willis is used as an alternative to invasive radiologic procedures and other MRI methods which require the application of contrast agents.
 We choose three different slides of this kind of MRI images with size $512\times 453$ and we provides the results of TFA and TFAE with the images which show different betwean this two results (Figure \ref{FIG04}). In the last column there are the images which show the vessele pixels which can be segmented by TFAE more than TFA (black pixels). In other word, black pixels can be distinguished as vessel by TFAE while they can not be segmented by TFA. \label{exam4}
\end{example}

\section{Conclusion }
In this paper, with contribution of the eigenvectors of Hessian matrix and the intensity value of pixels, we have
presented an algorithm for vessel extraction. The novelty stays on Stage 3 where a constraint is added to the thresholding
process. A slight change in the iteration part of TFA, actually shows that in the vessel segmentation has better result.
The experimental results illustrate how our algorithm does efficiently in tubular structure segmentation. Our algorithm
may be further improved in several ways. We choose Curvelets because of their ability to reconstruct curves. One can
choose other kind of wavelets family. Another way of improving the algorithm is to change the technique of computing
image gradient into other suitable PDE formulas. We choose reasonably (as described in Section 2) the Gaussian scale-space technique. Finally, we try to employ the Curvelet transform parameters as a default, one can optimize them.
\\

\bigskip

\noindent \textbf{BIOGRAPHY}\bigskip

\noindent \textbf{Nasser Aghazadeh} is an associate professor at \textit{Azarbaijan Shahid Madani University, Tabriz, Iran}. He obtained his
M.Sc. degree in Applied Mathematics (Numerical Analysis) from \textit{Iran University of Science \& Technology} and he
obtained his PhD degree in Applied Mathematics in the same place under supervision of \textit{Professor Khosrow Maleknejad}
on June 2007. His current research interests include numerical analysis, mathematical image processing, wavelets, etc.
He is a member of \textit{Research Group of Processing and Communication, Azarbaijan Shahid Madani University, Tabriz,
Iran.}  He can be contacted at aghazadeh@azaruniv.ac.ir.
\bigskip

\noindent \textbf{Ladan Sharafyan Cigaroudy} obtained her M.Sc. degree in Applied Mathematics from \textit{Tarbiat Modaress University of
Tehran, Iran} , in 2005. She is currently a PhD student in the Department of Applied Mathematics at \textit{Azarbaijan Shahid
Madani University}, Tabriz, Iran. She can be contacted at sharafyan@azaruniv.ac.ir.
\end{document}